\documentclass[conference]{IEEEtran}

\IEEEoverridecommandlockouts

\usepackage{cite}
\usepackage{amsmath,amssymb,amsfonts}
\usepackage{algorithmic}

\usepackage{graphicx}
\usepackage{textcomp}
\usepackage{xcolor}

\usepackage{array}
\usepackage{booktabs}
\usepackage{multirow}
\usepackage{tabularx}
\usepackage{makecell}

\usepackage{float}
\usepackage{stfloats}
\usepackage{placeins}
\usepackage{cuted}
\usepackage{capt-of}

\usepackage{fontspec}

\newfontfamily\bengalifont[
  Path=Fonts/,
  Extension=.ttf,
  Script=Bengali,
  Renderer=OpenType,
  BoldFont=NotoSerifBengali-Bold
]{NotoSerifBengali-Regular}

\newfontfamily\devanagarifont[
  Path=Fonts/,
  Extension=.ttf,
  Script=Devanagari,
  Renderer=OpenType,
  BoldFont=NotoSerifDevanagari-Bold
]{NotoSerifDevanagari-Regular}

\newfontfamily\urdufont[
  Path=Fonts/,
  Extension=.ttf,
  Script=Arabic,
  Renderer=OpenType,
  BoldFont=NotoSansArabic-Bold
]{NotoSansArabic-Regular}

\newfontfamily\tamilfont[
  Path=Fonts/,
  Extension=.ttf,
  Script=Tamil,
  Renderer=OpenType,
  BoldFont=NotoSerifTamil-Bold
]{NotoSerifTamil-Regular}

\newcommand{\textbengali}[1]{{\bengalifont #1}}
\newcommand{\texthindi}[1]{{\devanagarifont #1}}

\newcommand{\texturdu}[1]{{\urdufont #1}}

\newcommand{\texttamil}[1]{{\tamilfont #1}}
\newcommand{\nepalifont}{\devanagarifont}

\usepackage[hidelinks]{hyperref}

\newcommand{\twolinecell}[2]{%
  \begin{tabular}[t]{@{}p{\linewidth}@{}}%
    #1\\#2%
  \end{tabular}%
}

\def\BibTeX{{\rm B\kern-.05em{\sc i\kern-.025em b}\kern-.08em
    T\kern-.1667em\lower.7ex\hbox{E}\kern-.125emX}}
\begin{document}

\title{Analyzing performance of Retrieval Augmented Generation
models for Question Answering Tasks\\

}
\author{\IEEEauthorblockN{1\textsuperscript{st} Fazle Mohammed Tasfiq}
\IEEEauthorblockA{\textit{Dept. of Computer Science and Engineering} \\
\textit{BRAC University}\\
Dhaka, Bangladesh \\
fazle.mohammed.tasfiq@g.bracu.ac.bd}
\and
\IEEEauthorblockN{2\textsuperscript{nd} Sabab Attin}
\IEEEauthorblockA{\textit{Dept. of Computer Science and Engineering} \\
\textit{BRAC University}\\
Dhaka, Bangladesh \\
sabab.attin@g.bracu.ac.bd}
\and
\IEEEauthorblockN{3\textsuperscript{rd} Sheikh Md. Tahmid Hossain}
\IEEEauthorblockA{\textit{Dept. of Computer Science and Engineering} \\
\textit{BRAC University}\\
Dhaka, Bangladesh \\
sheikh.md.tahmidhossain@g.bracu.ac.bd}
\and
\IEEEauthorblockN{4\textsuperscript{th} Raiat Reza}
\IEEEauthorblockA{\textit{Dept. of Computer Science and Engineering} \\
\textit{BRAC University}\\
Dhaka, Bangladesh \\
raiat.reza@g.bracu.ac.bd}
\and
\IEEEauthorblockN{5\textsuperscript{th} Tanzim Ahmmed}
\IEEEauthorblockA{\textit{Dept. of Computer Science and Engineering} \\
\textit{BRAC University}\\
Dhaka, Bangladesh \\
tanzim.ahmmed@g.bracu.ac.bd}
\and
\IEEEauthorblockN{6\textsuperscript{th} Nafis Chowdhury}
\IEEEauthorblockA{\textit{Dept. of Computer Science and Engineering} \\
\textit{BRAC University}\\
Dhaka, Bangladesh \\
ext.nafis.chowdhury@bracu.ac.bd}
\and
\IEEEauthorblockN{7\textsuperscript{th} Anika Ahmed}
\IEEEauthorblockA{\textit{Dept. of Computer Science and Engineering} \\
\textit{BRAC University}\\
Dhaka, Bangladesh \\
anika.ahmed@bracu.ac.bd}
\and
\IEEEauthorblockN{8\textsuperscript{th} Moinul Haque}
\IEEEauthorblockA{\textit{Dept. of Computer Science and Engineering} \\
\textit{BRAC University}\\
Dhaka, Bangladesh \\
ext.moinul.haque@bracu.ac.bd}
\and[\\\hfill]
\IEEEauthorblockN{9\textsuperscript{th} Dr. Farig Yousuf Sadeque}
\IEEEauthorblockA{\textit{Dept. of Computer Science and Engineering} \\
\textit{BRAC University}\\
Dhaka, Bangladesh \\
farig.sadeque@bracu.ac.bd}
}

\maketitle

\begin{abstract}
Retrieval-Augmented Generation (RAG) models have become almost a cornerstone of open-domain question answering tasks; in which the quality of answers generated in the process, greatly relies on the effectiveness of retrieval mechanisms. In the realm of LLM, the last two years have seen tremendous progress in RAG architectures, with many models reporting to have improved performance through various design innovations. The cultural question and answering systems of South Asia, in particular, present a specific challenge since the region is home to the vast number of languages and traditions, which are closely interrelated. Along with these deeply intertwined historical narratives increase the need for culturally grounded and context-aware interpretation. Despite major architectural advances in LLMs, claiming significant improvements, challenges remain when answering these region specific questions in native languages. In this study, we aim to extract high-performing factors to guide the development of optimized hybrid RAG systems. We further explore whether the integration of RAG changes performance differently for culturally grounded and non-cultural questions in a South Asian multilingual setup, and whether RAG produces a higher overall percentage of correct answers than contexted answering. We also examine how LLM performance changes when RAG methods are integrated into the question-answering pipeline in comparison with non-contexted and contexted answering and, evaluate whether the introduction of RAG benefits popular or non-popular culturally grounded questions more. Not only does our work evaluate the current methods but also carves a path for future designs of more accurate and efficient question-answering pipelines.
\end{abstract}


\section{Introduction}
The rapid advancements in large language models (LLMs) have inaugurated a new era of natural language processing, enabling unprecedented capabilities in complex tasks such as summarization and Question Answering (QA). These models acquire their powerful, general-purpose language understanding through a process controlled by scaling laws, involving the training of billions of parameters on massive textual corpora \cite{minaee2024llmsurvey}. Prominent LLM families, including GPT, LLaMA, and PaLM, demonstrate this property by internalizing a vast set of parametric knowledge within their internal weights.
Yet, the operational utility of these systems for knowledge-intensive applications is fundamentally constrained by the static nature of their training data \cite{brown2025slr}. This structural limitation creates a defined knowledge cutoff, rendering models temporally bound and incapable of accessing or processing information created subsequent to their final training iteration. More critically, LLMs are systemically prone to hallucination, which is the generation of false, misleading, or unfounded information \cite{amirshahi2025robustness}. This phenomenon is often classified as intrinsic hallucination, where the output contradicts the immediate input context, or extrinsic hallucination, which arises when the model attempts to fill internal knowledge gaps by fabricating details inconsistent with its training data \cite{bang2025hallulens}. For factually demanding applications like QA, these inherent limitations severely compromise system reliability and diminish user trust.  
To overcome these fundamental barriers and augment LLMs with access to real-time, verifiable, and vastly expanded knowledge, the paradigm of Retrieval-Augmented Generation (RAG) was introduced in 2020 by Lewis et al \cite{lewis2020rag}. RAG fundamentally transforms the LLM architecture by fusing its pre-trained parametric memory with a dedicated non-parametric memory, an external, updatable knowledge base. This technique enables the LLM to dynamically retrieve relevant evidence from a vast corpus, conditioning its generation on verifiable, external context \cite{cheng2025knowledgeoriented}. By explicitly grounding its responses, RAG effectively expands the model’s factual scope beyond its original training boundary and significantly mitigates the risk of extrinsic hallucination. Consequently, RAG has rapidly become a central focus for research, proving capable of converting a general LLM into a more accurate, knowledgeable, and reliable system, particularly for complex QA tasks. 
The adoption of RAG models offers substantial benefits over traditional parametric LLMs, especially by generating language that is more specific, diverse, and factual compared to state-of-the-art parametric-only sequence-to-sequence baselines \cite{lewis2020rag}. Moreover, RAG is a strategic and efficient way of specialization. The ability to create and leverage a knowledge graph of external knowledge sources enables RAG to effectively transform a general LLM into a domain-specific expert \cite{wu2024ragsurvey}. This domain adaptation ability is a great benefit in terms of resource efficiency over the intensive and expensive fine-tuning that is needed to add equivalent specialized knowledge as parameters in the model. RAG has spread quickly, sparking an unprecedented surge of architectural research that has produced a new field of architecture, with a wide array of methods. Currently, RAG systems are widely categorized according to the innovations that are made in the retriever—how documents are selected, the generator—how context is synthesized, or hybrid approaches that combine coordinated innovations in both their respective components \cite{sharma2025survey}. This multi-component system, however, does have inherent complexity and potential areas for failure. Some of the basic tasks of RAG, including chunking large bodies of knowledge, embedding them in vectors, and indexing them efficiently in a database, are vital \cite{wu2024ragsurvey}. Even with the most powerful LLM, if these initial processes are not correct, the retriever will provide irrelevant or insufficient context. Thus, the analysis of performance for tasks in QA should go beyond linguistic fluency. Rather, evaluation requires specialized, diagnostic measures for factual consistency and verifiability. They include Groundedness (or Faithfulness): Does the response keep to the provided context? Context Relevance: Are the retrieved documents relevant to the question? Answer Relevance: How well does the final response answer the user's question? \cite{roychowdhury2024metrics} This will be a systematic assessment that will help to record progress and pinpoint the most successful RAG combinations. Now it is time to examine the performance dynamics of Retrieval-Augmented Generation models for QA tasks systematically, by looking at the influence of these design decisions and assessing their success by applying challenging, RAG-specific evaluations. This analysis can yield an actionable insight into building strong, reliable, and successful knowledge-empowered language models.

\begin{figure}[H]
\centering
\includegraphics[width=\columnwidth,height=0.38\textheight,keepaspectratio]{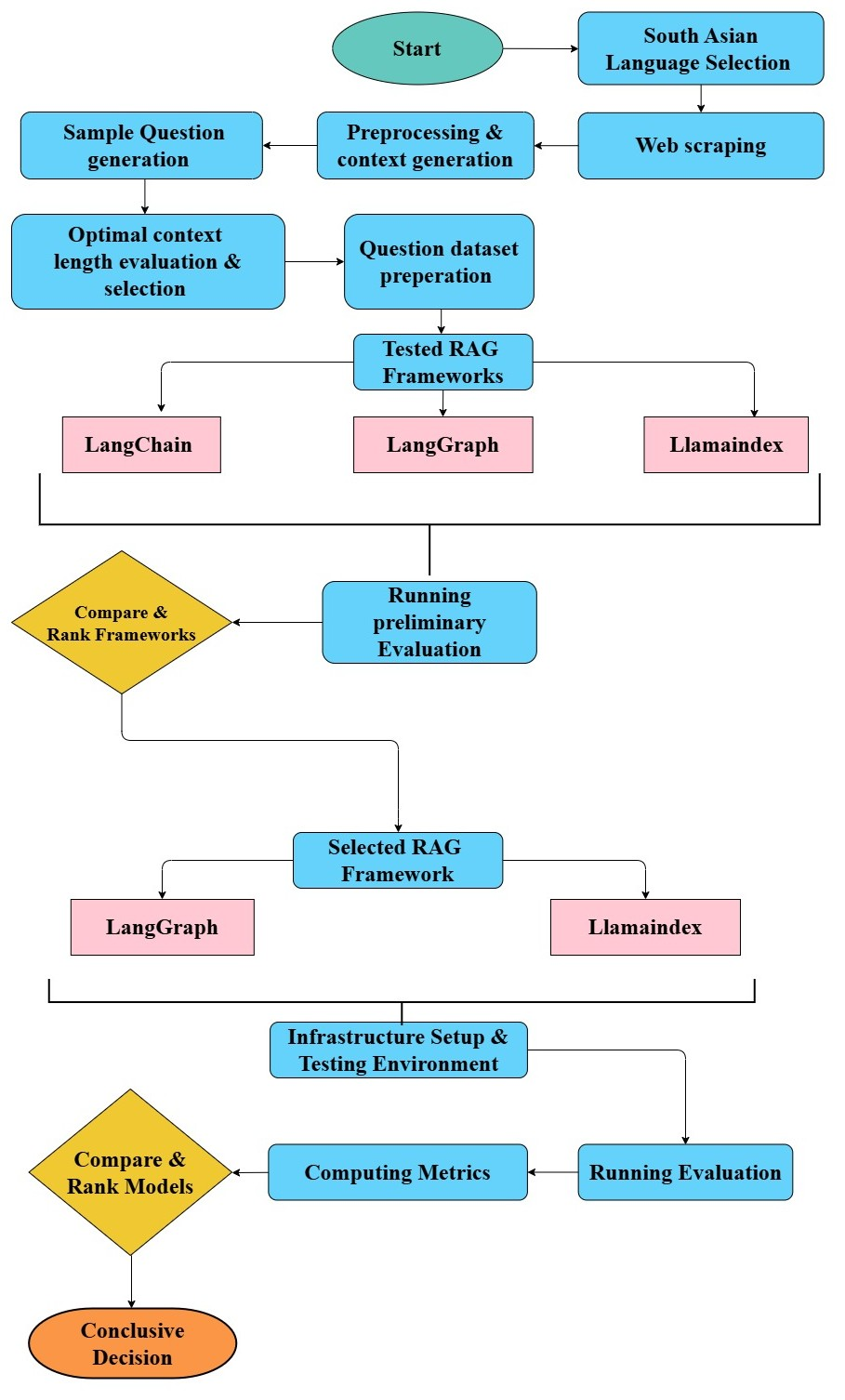}
\caption{Workflow of the RAG model evaluation framework.}
\label{fig:workflow_full}
\end{figure}

\section{Related Works}
Based on the overall finding, the assessment of RAG models in Question Answering(QA) needs to be expanded to beyond retrieval coverage. Rather, the three dimensions of structural knowledge management, execution efficiency and strong verification mechanisms should also be emphasized. The CRAG benchmark is the benchmark most commonly used as a performance benchmark, which quantifies the persistent Trust Gap, indicating that, in complex real-world QA settings, state-of-the-art RAG systems only achieve approximately 63\% truthfulness without hallucination \cite{yang2024crag}. The need for this has been met by recent advances in the architecture, which repeatedly show that structured knowledge fares much better than naive vector matching for complex queries. To illustrate, GraphRAG outperformed vector RAG in comprehensiveness with up to 83\% accuracy when it comes to the global contextual understanding of unstructured data as Knowledge Graphs (KGs) \cite{edge2024local}. This structural precision is realized in a useful and domain-specific manner by DO-RAG, which combines KG traversal with vector retrieval, resulting in the joint score gains of up to 33.38\% in technical domains \cite{opoku2025rag}. The efficiency challenge in dynamic corpora is solved by EraRAG \cite{zhang2025erarag}, whose LSH-based incremental update mechanism slashes KG reconstruction time by 77.5\%, validating that efficient structured storage is essential for modern RAG performance.\\\\
The second major pillar is performance optimization through adaptive and efficient execution.  On more complex multi-hop queries, Collab-RAG \cite{xu2025collab} proposes a breakthrough in query decomposition performance by training a small language model (SLM) to break down queries, showing that a 3B SLM can outperform a frozen 32B LLM in query decomposition performance. This highlights the importance of specialised, aligned orchestration models reducing the cost of running queries significantly \cite{xu2025collab}.  The achievable data and results of ARAGOG \cite{eibich2024aragog} indicate that the maximum benefits of optimisation of certain elements like the use of Hypothetical Document Embedding (HyDE) and reranking with LLM provide the highest benefits to the retrieval accuracy. This therefore confirms that the RAG performance is extremely sensitive to the specific combination and optimisation of high-impact elements.

\section{Dataset and Framework}
\subsection{Wikipedia webscraping}
The dataset construction process began with the selection of a subset of South Asian languages. Our choices included our native Bangla, along with Hindi, Urdu, Tamil and Nepali, covering the wider range of South Asia. However, since the languages we are focusing on are low-resource languages so there are no appropiate datasets for our experiments since most of the low-resource datasets are machine translated \cite{mahfuz2025too}. Upon selecting the languages, our task was to find appropriate datasets for these languages. The main motivation for constructing this multilingual dataset is inspired from the prior work by \cite{chowdhury2025facts}. To get the dataset for those languages, we needed accurate regional data. So we webscraped from the regional wikipedia pages including: \href{https://bn.wikipedia.org}{bn.wikipedia.org}, \href{https://ur.wikipedia.org}{ur.wikipedia.org}, \href{https://hi.wikipedia.org}{hi.wikipedia.org}, \href{https://ta.wikipedia.org}{ta.wikipedia.org}
and finally \href{https://ne.wikipedia.org}{ne.wikipedia.org} for native articles on Bangla, Urdu, Hindi, Tamil and Nepali respectively.  Our approach began by selecting the Wikipedia article of the country or state with the highest number of native speakers for a given language. This article served as the level-0 (landing) page. From this page, we followed in-page hyperlinks that redirected to separate Wikipedia articles, which were treated as level-1 terms. The process was then repeated for links found within level-1 pages to obtain level-2 terms. In total, this resulted in a three-level hierarchy of terms, comprising the landing page and two successive levels of hyperlink-based expansions. The levels were noted and to identify the common redirects, for each term a parent list was kept so that duplicate redirects can be avoided. In order to facilitate our tracking of the terms, all the corresponding URLs redirecting to the term were kept as well.

After collecting the titles and their corresponding article links, we observed a large number of Wikipedia page edit links as well as duplicate titles. Links containing \texttt{edit\&redlinks=1} were filtered out along with duplicate entries, and were removed. Additionally, some links redirected to Wikipedia calendar date pages, which were also discarded. Furthermore, stopwords were eliminated from our datasets with the help of publicly available stopword lists in kaggle and/or hugging face. 

In this way, for all the languages, the amount of data present in each dataset before and after preprocessing:

\begin{table}[H]
\centering
\caption{Dataset size before and after preprocessing}
\label{tab:preprocessing_stats}
\renewcommand{\arraystretch}{1.15}
\setlength{\tabcolsep}{3pt}
\footnotesize
\begin{tabular}{
    >{\centering\arraybackslash}p{0.25\columnwidth}
    >{\centering\arraybackslash}p{0.31\columnwidth}
    >{\centering\arraybackslash}p{0.31\columnwidth}
}
\toprule
\textbf{Language} & \textbf{Direct webscraping} & \textbf{After preprocessing} \\
\midrule
Bangla & 17,794 & 10,769 \\
Hindi & 16,362 & 9,541 \\
Tamil & 13,258 & 11,310 \\
Urdu & 8,795 & 7,805 \\
Nepali & 2,641 & 2,574 \\
\bottomrule
\end{tabular}
\end{table}

\subsection{Context generation}
Context was collected from Wikipedia for question generation. Our approach was that we used the links of the wikipedia pages from the database and scraped the first four lines from body paragraph content as context. However, in some cases, some pages did not contain a total of 4 lines, rather less. Some pages, specifically known as redlinks, had no content or included just a list of items and no body paragraph content. Contexts for such pages came up empty or zero. Along with this, we used Wikipedia Pageviewer API to list down the page view counts of each title in the time period from start to end of 2025 and normalized it between scores of 0 and 1 (otherwise the gap between counts are too sparse). Finally, the contents were taken and a counter variable was kept to note the number of context lines available. The generated files now had the title, url, context and number of lines and normalized popularity count as present in the sample table below :

\subsection{Question dataset construction}
The context we extracted was then used for question generation. Qwen 3 32b has been used to generate the question. We designed the prompts in  such a way that the model would design the question from the context given the answer of the question should be the term we collected and the term should not be present in the question. Two lines of context were used to generate the questions. The questions were generated by giving the same prompt translated to the other regional languages to keep consistency. 
The prompts for all the respective languages are translated into respective languages from this prompt-
“From the given context '\{context\}', write exactly one {name of language} question whose answer would be only '\{title\}'. The answer itself, or any part of it, must not be used in the question. There is no need to write anything other than the complete question that is generated.”
During the experimental phase, a problem arose that all the terms did not have 4 line contexts. But in most cases around 2 lines of context were available. We used 2 lines as context to retain maximum dataset and discarded those titles with only 1 line of context. Another case was observed that impacted this decision was the tendency of model hallucination when feeding too little (1 line) context or too much (4 line) context. 1 line context questions were almost too vague and 4 line context provided additional information for the model.

\subsection{Assessment of Question clarity}
A test sample of 250 questions on both 1-line and 2-line were analysed to determine how many lines of context provided better question. We used the following factors to evaluate: 

\textbf{Relevance:} The question correctly targets the core ideas present in the given answers. 

\textbf{Factual Accuracy:} The question does not introduce incorrect assumptions or false premises from the answers.

\textbf{Question Clarity:} The question is clearly and unambiguously phrased based on the answers.

\textbf{Creativity:} The question frames the existing answers in a novel, insightful, or non-trivial way. 

\textbf{Coherence:} The question is logically structured and internally consistent with the information in the answers.

\textbf{Task-Specific Appropriateness:} title not repeated in question, question has only the complete title as answer, no other answer is possible except the title and question is in expected language or not. 

\vspace{0.5cm}
All the ratings were done by manual human input out of 5, and then an average was calculated. Upon the evaluation, it was found out that in case of all languages, better questions were observed for 2 line contexts. In case of other languages, the respective questions were translated to Bangla first via google translator model call and then evaluated. Since these questions were all judged on the basis of translation, we could not judge the coherence. So no coherence evaluation was done and the values might vary if judged by a native speaker.
It was found that the performance is better when there are 2 lines of context used while question generation. After the confirmation of using 2 line contexts, we proceeded towards full set question generation and answer generation. All the questions were generated by the same prompt as mentioned before. For answer generation, we used four models, 3 open and 1 closed source for diversity. Llama 4 Maverick 17B by Meta (open source), DeepSeek V3-0324 (open source), Gemma 3 27B by Google (open source) and Grok 4.1 Fast by xAI (closed source).

\subsection{LLM-Based Semantic Classification}
We followed an LLM-assisted semantic classification method. The categorization scheme from Chowdhury et al.’s \cite{chowdhury2025facts} Bengali cultural knowledge evaluation provided a peer-reviewed taxonomy for organizing titles into meaningful semantic groups that worked as a reference structure for this stage. Each title was treated as an individual instance and passed to Meta Llama Maverick with a well-constrained instruction to return only one label from the predefined set of labels. This LLM was selected upon evaluating its stronger performance in question-answering tasks in our prior analysis. The model outputs were then stored as semantic labels for each title.

The following definitions were provided in the model instructions when passing the titles:

\newpage
\begin{strip}
\centering
\refstepcounter{table}
\label{tab:categories}
{\scshape Table \Roman{table}}\\[2pt]
{\footnotesize Category definitions, classes, and examples}\\[6pt]
\renewcommand{\arraystretch}{1.12}
\setlength{\tabcolsep}{3pt}
\scriptsize
\begin{tabularx}{\textwidth}{
    >{\raggedright\arraybackslash}p{0.12\textwidth}
    >{\centering\arraybackslash}p{0.11\textwidth}
    >{\raggedright\arraybackslash}X
    >{\raggedright\arraybackslash}p{0.25\textwidth}
}
\toprule
\textbf{Category} & \textbf{Class} & \textbf{Definition} & \textbf{Example} \\
\midrule
Event & Cultural & Specific occurrences such as festivals, battles, conferences, natural disasters, ceremonies, or incidents. & \makecell[l]{\textbengali{পহেলা বৈশাখ}\\{}[Pohela Boishakh]} \\
\hline
Religion & Cultural & Religious beliefs, practices, deities, scriptures, rituals, or religious movements. & \makecell[l]{\textbengali{বৌদ্ধধর্ম}\\{}[Buddhism]} \\
\hline
Food & Cultural & Dishes, ingredients, cuisines, beverages, or culinary items. & \makecell[l]{\textbengali{রসগোল্লা}\\{}[Rasgulla]} \\
\hline
Entertainment & Cultural & Movies, music, television, games, arts, celebrities, or media-related content. & \makecell[l]{\textbengali{লোকসঙ্গীত}\\{}[Folk music]} \\
\hline
Historical & Cultural & Historical eras, dynasties, treaties, or historically significant topics not classified as specific events. & \makecell[l]{\textbengali{গুপ্ত সাম্রাজ্য}\\{}[Gupta Empire]} \\
\hline
People & Cultural & Individual persons, real or fictional, including public figures, scholars, artists, and athletes. & \makecell[l]{\textbengali{কাজী নজরুল ইসলাম}\\{}[Kazi Nazrul Islam]} \\
\hline
Politics & Non-cultural & Political parties, ideologies, political offices, public policies, governance, or political movements. & \makecell[l]{\textbengali{বাংলাদেশের সংবিধান}\\{}[Constitution of Bangladesh]} \\
\hline
Locations & Non-cultural & Countries, cities, regions, landmarks, geographical entities, or physical places. & \makecell[l]{\textbengali{বাংলাদেশ}\\{}[Bangladesh]} \\
\hline
Language & Non-cultural & Languages, dialects, scripts, linguistic concepts, or grammar-related topics. & \makecell[l]{\textbengali{বাংলা ভাষা}\\{}[Bangla Language]} \\
\hline
Material & Non-cultural & Physical substances, chemicals, elements, minerals, fabrics, or tangible materials. & \makecell[l]{\textbengali{মসলিন}\\{}[Muslin]} \\
\hline
Organization & Non-cultural & Companies, institutions, NGOs, agencies, universities, clubs, or structured groups. & \makecell[l]{\textbengali{দৈনিক প্রথম আলো}\\{}[The Daily Prothom Alo]} \\
\hline
Health & Non-cultural & Diseases, medical conditions, treatments, anatomy, healthcare systems, or psychological conditions. & \makecell[l]{\textbengali{ডেঙ্গু জ্বর}\\{}[Dengue Fever]} \\
\hline
Temporal & Non-cultural & Time-related concepts such as months, years, calendars, historical periods when referring to time itself, or dates. & \makecell[l]{\textbengali{চৈত্র}\\{}[Chaitra]} \\
\hline
Economics & Non-cultural & Financial systems, markets, economic theories, currencies, trade, or economic indicators. & \makecell[l]{\textbengali{বাংলাদেশের অর্থনীতি}\\{}[Economy of Bangladesh]} \\
\hline
Sports & Non-cultural & Sports types, tournaments, teams, competitions, or athletic disciplines. & \makecell[l]{\textbengali{ক্রিকেট}\\{}[Cricket]} \\
\hline
Miscellaneous & Non-cultural & Titles that do not clearly belong to any of the above categories. & \makecell[l]{\textbengali{রিকশা}\\{}[Rickshaw]} \\
\bottomrule
\end{tabularx}
\end{strip}

\subsection{Dataset Analysis}

After the classification was done, the full dataset was formulated and, since we pivoted to 2 line context, the dataset size reduced from the previous ones of~\autoref{tab:preprocessing_stats} to:

\begin{table}[!t]
\centering
\caption{Amount of data present in the final dataset}
\label{tab:qa_preprocessing_stats}
\renewcommand{\arraystretch}{1.1}
\setlength{\tabcolsep}{6pt}
\footnotesize
\begin{tabularx}{0.88\columnwidth}{
    >{\centering\arraybackslash}X
    >{\centering\arraybackslash}X
}
\toprule
\textbf{Language} & \textbf{Final data count} \\
\midrule
Bangla & 10,278 \\
Hindi & 8,694 \\
Tamil & 8,553 \\
Urdu & 5,806 \\
Nepali & 2,405 \\
\bottomrule
\end{tabularx}
\end{table}

Our final dataset hence had 4 columns: Title, context 2 line, question 2 line and cluster i.e. the categories. The dataset is available at \cite{tasfiq2026sacmq}.

\begin{table*}[!t]
\centering
\caption{Sample of multilingual QA dataset}
\label{tab:bangla_qa_sample}
\renewcommand{\arraystretch}{1.12}
\setlength{\tabcolsep}{3pt}
\scriptsize
\begin{tabularx}{\textwidth}{
    >{\raggedright\arraybackslash}p{0.14\textwidth}
    >{\raggedright\arraybackslash}X
    >{\raggedright\arraybackslash}X
    >{\centering\arraybackslash}p{0.10\textwidth}
}
\toprule
\textbf{Title} & \textbf{Context} & \textbf{Question} & \textbf{Cluster} \\
\midrule
\twolinecell{\textbengali{দৈনিক ইত্তেফাক}}{The Daily Ittefaq} &
\twolinecell{\textbengali{দৈনিক ইত্তেফাক বাংলাদেশের বাংলা ভাষার একটি জাতীয় দৈনিক সংবাদপত্র। এটি বাংলা ভাষার প্রিন্ট সংস্করণ ছাড়াও ইংরেজি অনলাইন সংস্করণে প্রকাশিত হয়।}}{The Daily Ittefaq is a national daily newspaper published in the Bengali language in Bangladesh. It also has an English online edition.} &
\twolinecell{\textbengali{বাংলাদেশের বাংলা ভাষার একটি জাতীয় দৈনিক সংবাদপত্র কী?}}{What is a national daily newspaper in the Bengali language of Bangladesh?} &
organization \\
\midrule
\twolinecell{\texthindi{मुज़फ़्फ़रनगर}}{Muzaffarnagar} &
\twolinecell{\texthindi{मुज़फ़्फ़रनगर भारत के उत्तर प्रदेश राज्य के मुज़फ़्फ़रनगर ज़िले में स्थित एक नगर है। मुज़फ़्फ़रनगर यह ज़िले का मुख्यालय भी है।}}{Muzaffarnagar is a city in Uttar Pradesh, India, and the headquarters of Muzaffarnagar district.} &
\twolinecell{\texthindi{उत्तर प्रदेश के किस जिले का मुख्यालय उसी नाम के शहर में स्थित है?}}{Which district of Uttar Pradesh has its headquarters in a city with the same name?} &
locations \\
\midrule
\twolinecell{\texturdu{کشمیری زبان}}{Kashmiri language} &
\twolinecell{\texturdu{کشمیری زبان بھارت اور پاکستان کی ایک اہم زبان ہے۔ متکلمین کی تعداد 15 لاکھ سے کچھ اوپر ہے۔}}{The Kashmiri language is an important language of India and Pakistan, with slightly over 1.5 million speakers.} &
\twolinecell{\texturdu{بھارت اور پاکستان میں بولی جانے والی اس زبان کا نام کیا ہے جس کے تقریباً 15 لاکھ سے زائد بولنے والے ہیں؟}}{What language is spoken in India and Pakistan by over 1.5 million people?} &
language \\
\midrule
\twolinecell{\texttamil{பல்லாங்குழி}}{Pallankuzhi} &
\twolinecell{\texttamil{பல்லாங்குழி என்பது, பொதுவாக பெண்கள் விளையாடும் ஒரு விளையாட்டு. முதற் பூப்படைந்த பெண்ணின் தீட்டுக்குரிய காலத்திலும் கருவுற்ற பெண்கள் அமர்ந்து பொழுதுபோக்குவதற்கும் இந்த விளையாட்டை இப்பொழுது ஆடிப் பார்க்கிறார்கள்.}}{Pallanguzhi is a traditional game generally played by women.} &
\twolinecell{\texttamil{பெண்கள் விளையாடும் பழமைவாய்ந்த விளையாட்டு எது?}}{Which traditional game is played by women?} &
entertainment \\
\midrule
\twolinecell{{\nepalifont हिमालय}}{Himalayas} &
\twolinecell{{\nepalifont हिमालय एसिया को एक हिमशृङ्खला हो, जसले भारतीय उपमहाद्वीप को मैदानलाई तिब्बती पठार बाट अलग गर्छ। यस हिमशृङ्खलामा पृथ्वी का केही अग्लो चुचुराहरू छन्, जसमा सबैभन्दा अग्लो, सगरमाथा पनि समावेश छ।}}{The Himalayas separate the Indian subcontinent from the Tibetan Plateau and include Mount Everest.} &
\twolinecell{{\nepalifont भारतीय उपमहाद्वीपको मैदानलाई तिब्बती पठारबाट अलग पार्ने एवं सगरमाथाले समेटिएको एसियाको एउटा हिमशृङ्खला के हो?}}{What mountain range separates the Indian subcontinent from the Tibetan Plateau and includes Mount Everest?} &
locations \\
\bottomrule
\end{tabularx}
\end{table*}

\FloatBarrier

\section{RAG Setup and Analysis}
The next part of our work adopted Retrieval-Augmented Generation as the main framework for context-supported answer generation. We integrated RAG to our pipeline because it combines a parametric LLM with a non-parametric retrievable memory base, thus allowing generation to be constrained on relevant retrieved contexts or ground truths instead of only relying on the model’s base knowledge \cite{lewis2020rag} . In our setup, the retrieved context served as the external knowledge base or ground truth, while the LLM generated the final answer from the query and retrieved passages.
This phase began with selecting three frameworks for implementation of the pilot evaluation of RAG-based answer generation- LangChain, LangGraph, and LlamaIndex. Each of these frameworks provides a different approach towards organizing retrieval-based generation. LangChain was included because it supports chain-based composition of LLM workflows, where retrieval, prompting, model invocation, and output parsing can be connected sequentially \cite{chase2022langchain} . In a slightly different approach, LangGraph represents applications through a shared state of nodes and edges to update the state and define next execution steps \cite{langgraph2024} . LlamaIndex provides retrieval-augmented LLM applications with data indices and query engines, where each engine provides natural-language access to indexed data \cite{Liu_LlamaIndex_2022} . 
A 100-sample pilot test was carried out across the three frameworks under similar retrieval and generation settings for each language using each LLM. The results showed weaker performance of LangChain based pipeline compared to the other two, specifically for lower performing language settings such as Nepali, Tamil, and Urdu. As a result, further work was carried out using LangGraph and LlamaIndex as RAG frameworks.

\begin{table}[H]
\centering
\caption{Performance comparison across languages and frameworks}
\label{tab:llm_performance}
\renewcommand{\arraystretch}{0.95}
\scriptsize
\begin{tabular*}{\columnwidth}{@{\extracolsep{\fill}}llccc@{}}
\toprule
\textbf{Lang.} & \textbf{LLM} & \textbf{LC} & \textbf{LG} & \textbf{LI} \\
\midrule
\multirow{4}{*}{Bangla}
& DeepSeek V3 & 64.0 & 64.0 & 57.0 \\
& Gemma 3 & 67.0 & 66.0 & 70.0 \\
& LLaMA 4 & 66.0 & 66.0 & 68.0 \\
& Grok 4.1 & 61.0 & 65.0 & 60.0 \\
\midrule
\multirow{4}{*}{Hindi}
& DeepSeek V3 & 37.0 & 38.0 & 34.0 \\
& Gemma 3 & 46.0 & 46.0 & 44.0 \\
& LLaMA 4 & 63.0 & 63.0 & 64.0 \\
& Grok 4.1 & 56.0 & 60.0 & 58.0 \\
\midrule
\multirow{4}{*}{Nepali}
& DeepSeek V3 & 51.0 & 53.0 & 55.0 \\
& Gemma 3 & 53.0 & 55.0 & 53.0 \\
& LLaMA 4 & 64.0 & 74.0 & 76.0 \\
& Grok 4.1 & 60.0 & 67.0 & 61.0 \\
\midrule
\multirow{4}{*}{Tamil}
& DeepSeek V3 & 40.0 & 43.0 & 41.0 \\
& Gemma 3 & 49.0 & 46.0 & 51.0 \\
& LLaMA 4 & 63.0 & 66.0 & 68.0 \\
& Grok 4.1 & 44.0 & 41.0 & 44.0 \\
\midrule
\multirow{4}{*}{Urdu}
& DeepSeek V3 & 38.0 & 39.0 & 42.0 \\
& Gemma 3 & 11.0 & 10.0 & 15.0 \\
& LLaMA 4 & 57.0 & 68.0 & 70.0 \\
& Grok 4.1 & 61.0 & 63.0 & 63.0 \\
\bottomrule
\end{tabular*}

\vspace{2pt}
\footnotesize
LC = LangChain, LG = LangGraph, LI = LlamaIndex. Values are percentages.
\end{table}

Throughout the rest of our work, LangGraph and LlamaIndex organized the same RAG task following different pipeline structures. LangGraph pipeline modeled the process as a state-based workflow to store retrieved contexts and generated answers as a part of a shared RAG state. The retrieval and generation were executed as separate nodes that were dependent directly on the context returned by the retriever. On the other hand, LlamaIndex connected the documents to a Chroma vector space and queried through a query-engine following an index-based approach. In both cases, the pipelines followed similar retrieval to generation logic, but the implementation differed in how context was retrieved, passed and used during answer generation processes.

\section{RAG Pipeline: ChromaDB, Embeddings, and Retrieval} 
Our RAG pipeline used ChromaDB as the vector store since the retrieval stage required semantic similarity search rather than conventional keyword lookup. ChromaDB stores document chunks as embeddings along with their corresponding documents and metadata making it suitable for RAG-based AI workflows where query embeddings must be matched with semantically relevant stored passages \cite{chroma2025}. In our pipeline, the query was embedded at runtime and compared against the stored chunk embeddings so that the most relevant contexts could be retrieved and passed as ground truth for answer generation.
The knowledge base was prepared from the \texttt{rag\_database.csv} file, where each entry contained a title and a short two-line contextual description. Each row was converted into a document by combining the title with its corresponding context before chunking and embedding. A sample of the dataset structure is shown below:

\vspace{2\baselineskip}
\begin{strip}
\centering
\refstepcounter{table}
\label{tab:bangla_titles_context}
{\scshape Table \Roman{table}}\\[2pt]
{\footnotesize Sample Bangla Titles with Contextual Descriptions}\\[6pt]
\renewcommand{\arraystretch}{1.25}
\setlength{\tabcolsep}{4pt}
\small
\begin{tabularx}{\textwidth}{
    >{\raggedright\arraybackslash}p{0.18\textwidth}
    >{\raggedright\arraybackslash}X
    >{\centering\arraybackslash}p{0.08\textwidth}
}
\toprule
\textbf{Title} & \textbf{Context} & \textbf{Language} \\
\midrule
\twolinecell{\textbengali{বাংলাদেশ}}{Bangladesh} &
\twolinecell{\textbengali{বাংলাদেশ দক্ষিণ এশিয়ার একটি স্বাধীন সার্বভৌম রাষ্ট্র। বাংলাদেশের সাংবিধানিক নাম গণপ্রজাতন্ত্রী বাংলাদেশ।}}{Bangladesh is an independent sovereign state in South Asia. The constitutional name of Bangladesh is the People's Republic of Bangladesh.} &
bn \\
\midrule
\twolinecell{\textbengali{ওসমানী আন্তর্জাতিক বিমানবন্দর}}{Osmani International Airport} &
\twolinecell{\textbengali{ওসমানী আন্তর্জাতিক বিমানবন্দর বাংলাদেশের অন্যতম প্রধান আন্তর্জাতিক বিমানবন্দর। এটি সিলেটে অবস্থিত।}}{Osmani International Airport is one of the major international airports in Bangladesh. It is located in Sylhet.} &
bn \\
\midrule
\twolinecell{\textbengali{দৈনিক প্রথম আলো}}{The Daily Prothom Alo} &
\twolinecell{\textbengali{দৈনিক প্রথম আলো বাংলাদেশ থেকে বাংলা ভাষায় প্রকাশিত একটি দৈনিক সংবাদপত্র। ১৯৯৮ সালের ৪ নভেম্বর এটি প্রথম প্রকাশিত হয়।}}{Prothom Alo is a daily newspaper published in Bengali from Bangladesh. It was first published on 4 November 1998.} &
bn \\
\bottomrule
\end{tabularx}
\end{strip}

The documents were further split into smaller chunks before embedding. For embedding generation, we used intfloat/multilingual-e5-large-instruct, a multilingual embedding model designed for text retrieval tasks and released as a part of the Multilingual-E5 embedding family \cite{wang2024multilinguale5} .
Separate ChromaDB directories were maintained for each framework implementation in order to maintain framework-specific indexing behavior independent of each other aligning with reproducible research practices \cite{sandve2013reproducible}.

\section{Result Collection}
\subsection{Initial Answer Generation}
Answer generation was done by api calls and the same prompt for generating answers were translated to their respective languages. In order to evaluate the baseline knowledge of the models in each language, and their ability to utilize contextual information and also evaluate the change in performance, we generate answers in two stages. At first, answers were generated without providing context to the models (non-contexted answer) and then by providing context (contexted answers). Similarly answer generation was done by api calls and the same prompt for generating answers were translated to their respective languages. In order to evaluate the baseline knowledge of the models in each language, and their ability to utilize contextual information and also evaluate the change in performance, we generate answers in two stages. At first, answers were generated without providing context to the models (non-contexted answer) and then by providing context (contexted answers). The prompts used in this case was in english as followed for Non-contexted answer generation.
“According to the question ‘\{question\}’, provide only the direct correct answer to the question. Try to limit the answer to one word, and under no circumstances should you use more than five words. Do not add any explanation, full sentence, punctuation, extra words, introduction, or conclusion—only the final answer should be given in the output.”
and for contexted answer generation the prompt was-
“According to the question ‘\{question\}’ and the given context ‘\{context\}’, provide only the direct correct answer to the question. Try to limit the answer to one word, and under no circumstances should you use more than five words. Do not add any explanation, full sentence, punctuation, extra words, introduction, or conclusion—only the final answer should be given in the output.”
these two prompts were translated into respective languages.

\subsection{RAG-Based Answer Collection}

RAG-based answer collection was conducted after the final selection of LangGraph and LlamaIndex. For each language, model, and framework combination, the generated questions were processed through the corresponding RAG pipeline, and the model outputs were saved for evaluation. The same answer-generation instruction was used across all settings, with language-specific translations to maintain consistency.

During the answer generation stage, the input question was embedded and matched against the corresponding ChromaDB vector store, and the top retrieved chunks were passed to the LLM along with the question. The prompt was designed to produce short title-like answers rather than explanatory responses. Since the questions were constructed so that the expected answer was the corresponding title, each model was instructed to return only the final answer. The collected outputs were stored with their language, model, framework, question, retrieved context, generated answer, and reference title. The following prompt was used in respective languages to generate RAG-Based Answer for each language-
"You are a \{Language\} cultural question-answering assistant. Answer the question from the context below. Give the answer in \{Language\}. Limit the answer to one word, and under no circumstances should you use more than five words; there is no need to write anything extra or a complete sentence. If the answer is not found in the context, say: “Information not found.” Context: \{context\} Question: \{question\}."

\section{Model}
\label{sec:Model}
\phantomsection\label{tab:ai_models_purpose}
\subsection{Question Generation}
Qwen 3 32B by Alibaba is used exclusively for question generation. With 32 billion parameters and support for over 100 languages, it operates in two modes: a structured "thinking" mode for carefully constructed prompts and a faster "non-thinking" mode for fluent question variations. Its 131,072-token context window enables diverse, context-aware question creation across topics and languages.
\subsection{Answer Generation}
Four models handle answer generation, each bringing distinct strengths. Meta Llama 4 Maverick 17B is a multimodal open-source model with 400 billion parameters and 128 experts, handling both text and image inputs with a 128K context window, excelling in natural language processing and long-context applications. DeepSeek Chat V3-0324 is a massive 685-billion-parameter mixture-of-experts model with a 163,840-token context window, built for high-performance natural language understanding and complex dialogues. Google Gemma 3 27B supports vision-language input and text output across 140+ languages, performing strongly in math, reasoning, and structured outputs with a 128K context window and function calling capabilities. xAI Grok 4.1 Fast is a closed-source agentic model featuring a remarkable 2-million-token context window, optimized for customer support and deep research with an optional reasoning mode for enhanced decision-making.
\subsection{Translation}
OpenAI GPT-OSS 120B is an open-weight, 117-billion-parameter mixture-of-experts model used solely for translation across five languages. With a 131K context window, native MXFP4 quantization for efficient inference, and support for function calling and structured outputs, it brings high-reasoning capacity to multilingual translation workflows.

\section{Result and Analysis}
\label{sec:dataset_and_analysis}
\subsection{Result Analysis}

\begin{figure}[H]
\centering
\includegraphics[width=\columnwidth]{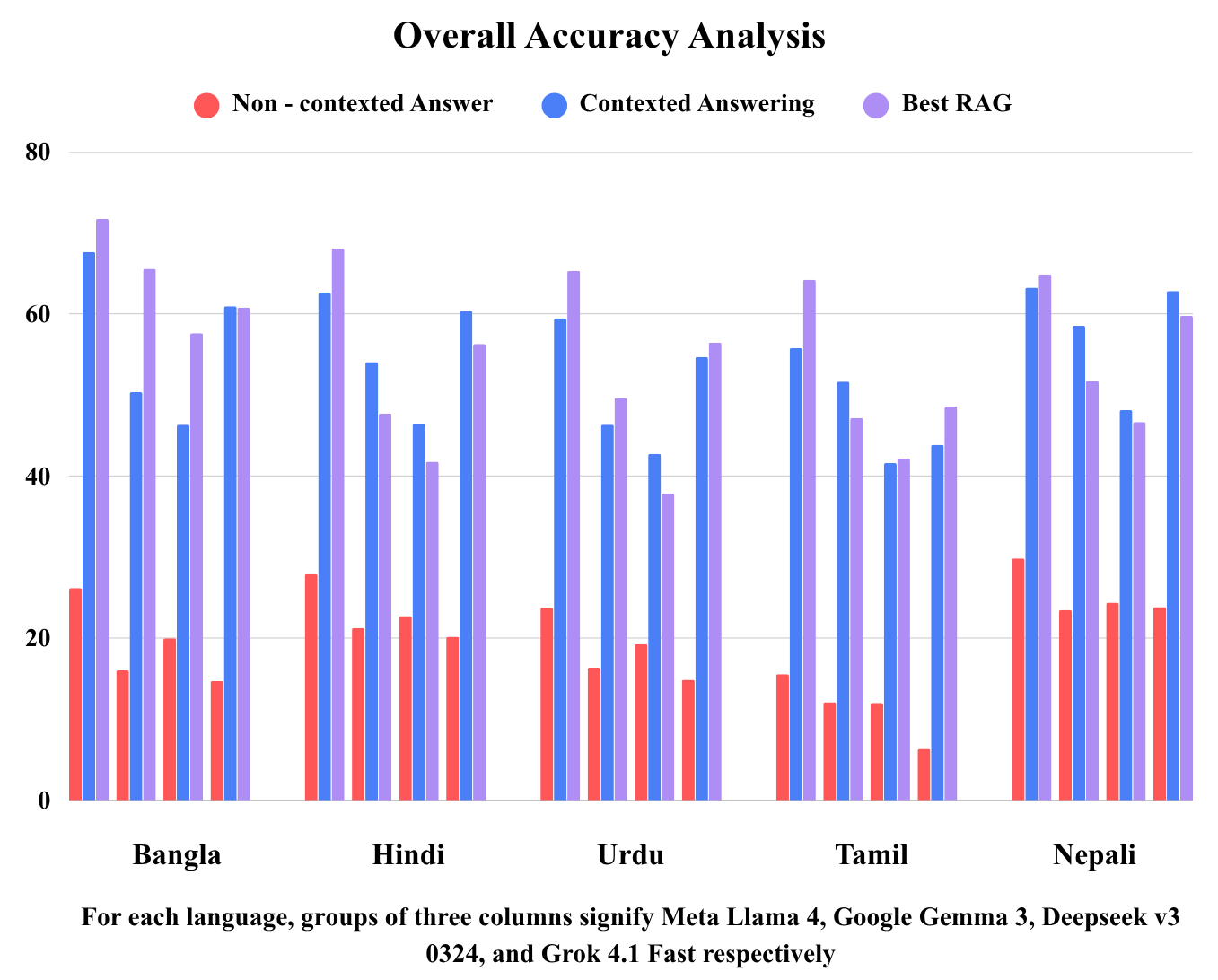}
\caption{Overall accuracy comparison across non-contexted, contexted, and best RAG settings for five languages. Model groups within each language represent Meta Llama 4, Google Gemma 3, DeepSeek V3-0324, and Grok 4.1 Fast, respectively.}
\label{fig:overall-accuracy-analysis}
\end{figure}

The overall accuracy results show a clear gap between non-contexted and context-supported answer generation. As shown in Fig.~\ref{fig:overall-accuracy-analysis}, both contexted answering and best RAG perform substantially better than non-contexted answering across all five languages. Meta Llama 4 Maverick 17B generally achieves the strongest performance across the evaluated settings, while DeepSeek V3-0324 and Grok 4.1 Fast show weaker performance depending on the language and answering condition. However, several language- and category-specific anomalies are observed. In Nepali, Google Gemma 3 27B performs weakest in the general non-contexted settings, while DeepSeek V3-0324 produces the weakest performance for overall non-popular titles and non-cultural non-popular titles under non-contexted answering. Another exception appears in culturally grounded non-popular contexted answering, where Grok 4.1 Fast performs better than Meta Llama 4 Maverick. Similar deviations are also visible in Hindi and Urdu. In Hindi cultural non-contexted answering, Google Gemma 3 27B performs weakest instead of Grok. In Urdu, Gemma performs weakest for popular culturally grounded non-contexted answering and popular non-cultural contexted answering.

\begin{figure}[H]
\centering
\includegraphics[width=\columnwidth]{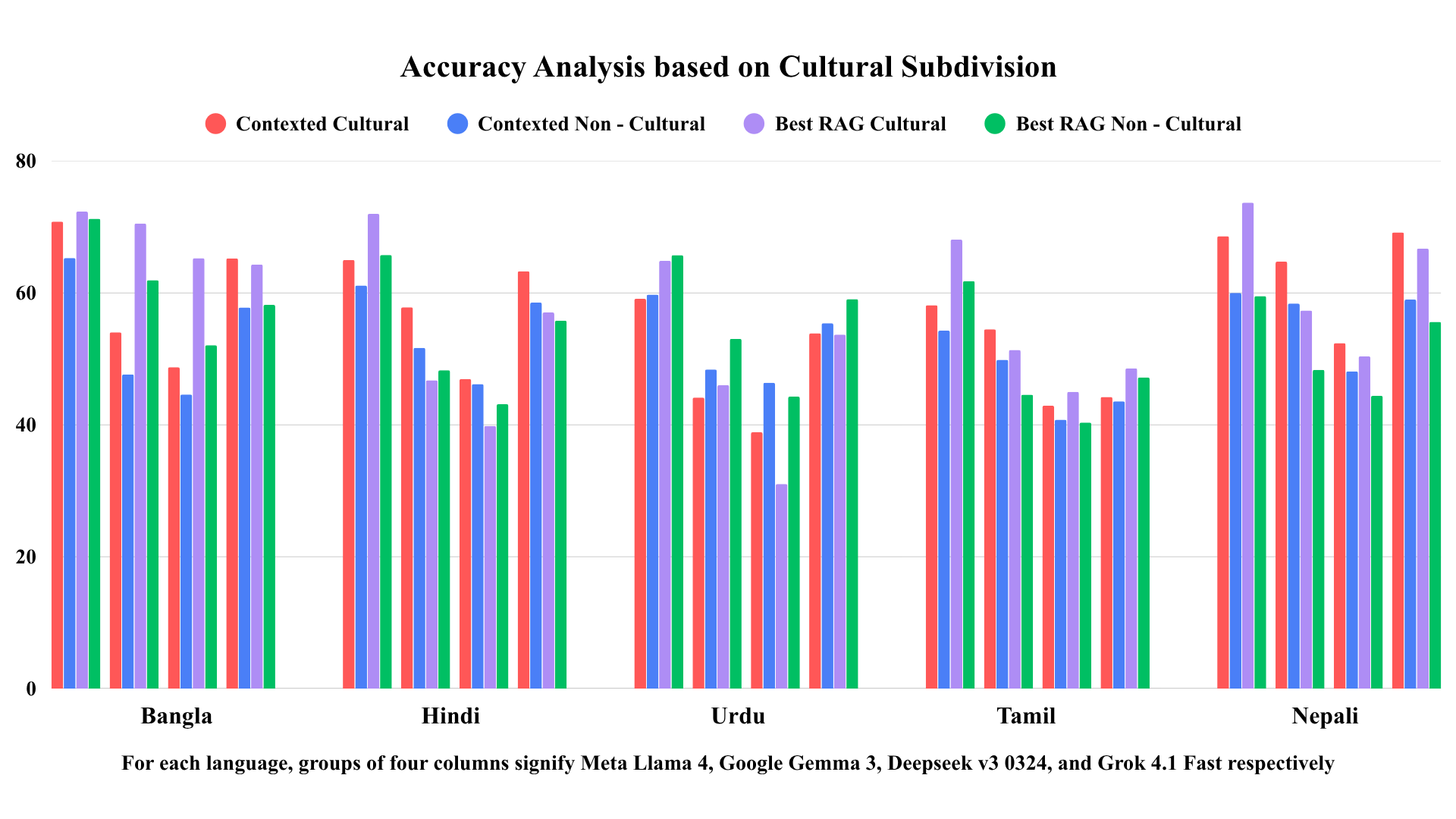}
\caption{Accuracy comparison across cultural and non-cultural question categories under contexted and best RAG settings for five languages. Model groups within each language represent Meta Llama 4, Google Gemma 3, DeepSeek V3-0324, and Grok 4.1 Fast, respectively.}
\label{fig:cultural-subdivision-analysis}
\end{figure}

The cultural subdivision results in Fig.~\ref{fig:cultural-subdivision-analysis} show that performance does not vary uniformly between cultural and non-cultural questions. In several languages, cultural questions achieve comparable or higher accuracy than non-cultural questions, while in other cases the opposite pattern is observed. This indicates that cultural relevance alone does not determine answer-generation accuracy. Instead, the results vary depending on the language, model, and type of context used. The figure also shows that RAG does not consistently improve both cultural and non-cultural categories for every model, which supports the need for language-wise and model-wise analysis.

\begin{figure}[H]
\centering
\includegraphics[width=\columnwidth]{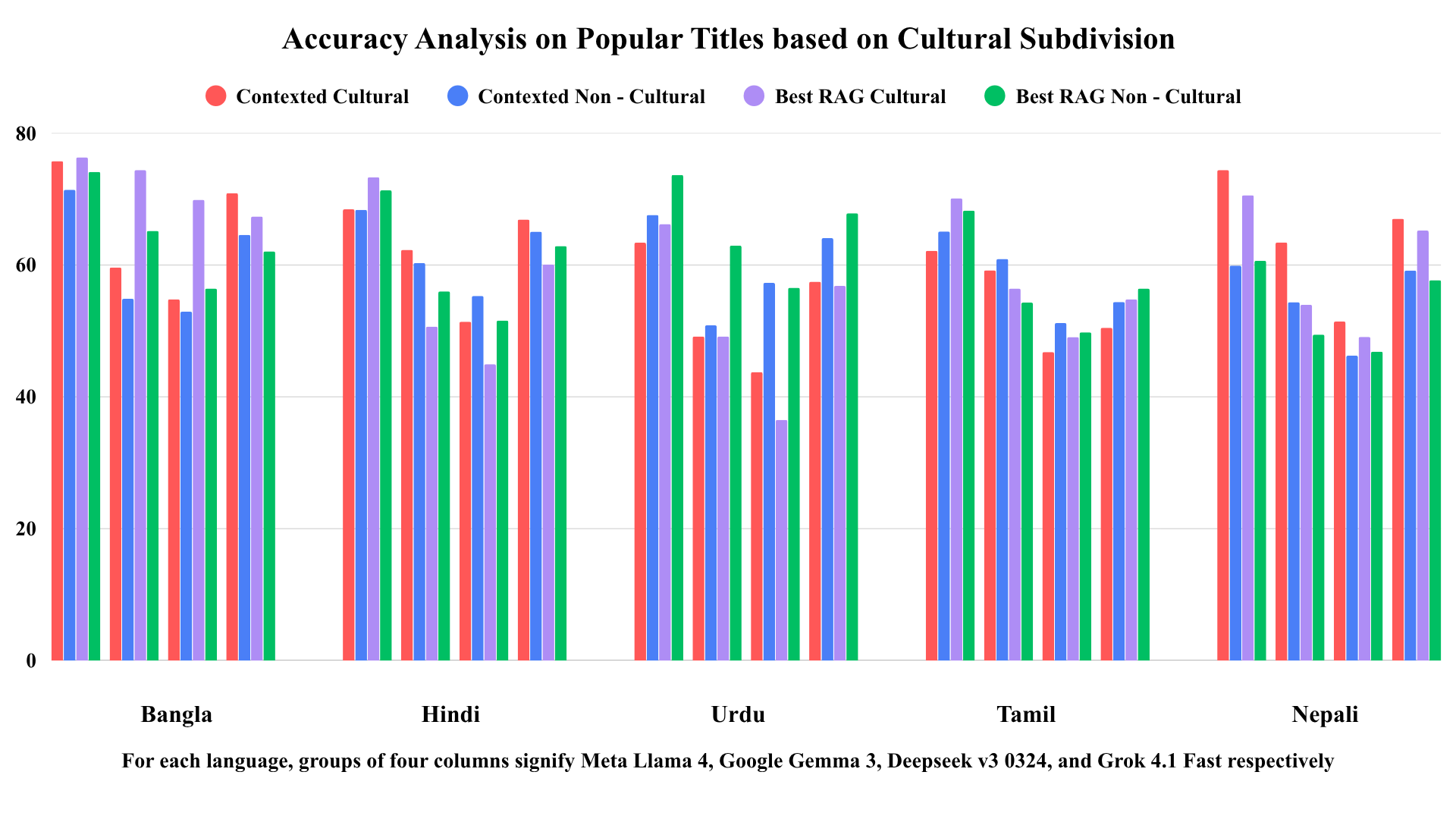}
\caption{Accuracy comparison for popular titles across cultural and non-cultural question categories under contexted and best RAG settings. Model groups within each language represent Meta Llama 4, Google Gemma 3, DeepSeek V3-0324, and Grok 4.1 Fast, respectively.}
\label{fig:popular-cultural-subdivision}
\end{figure}

\begin{figure}[H]
\centering
\includegraphics[width=\columnwidth]{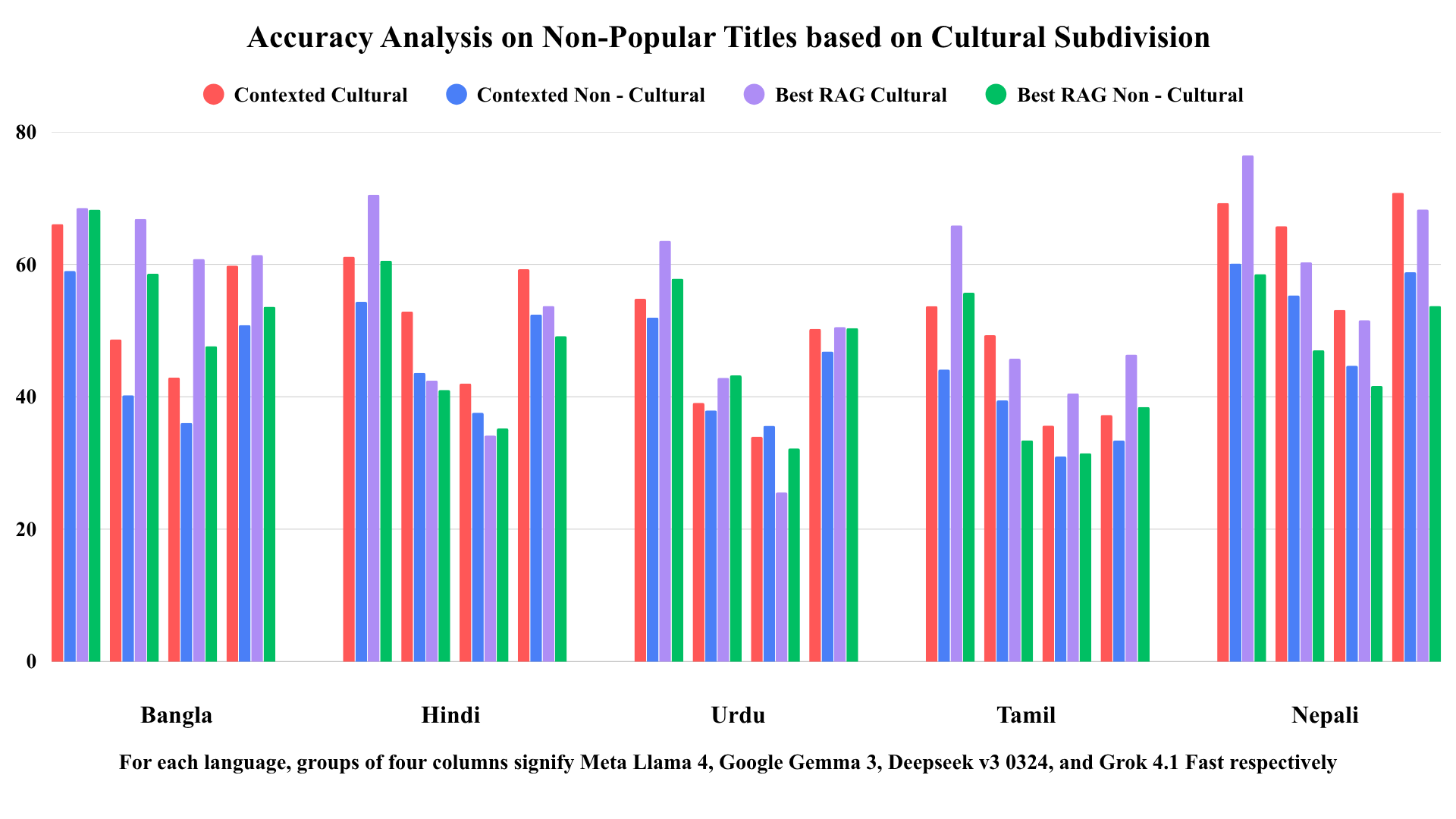}
\caption{Accuracy comparison for non-popular titles across cultural and non-cultural question categories under contexted and best RAG settings. Model groups within each language represent Meta Llama 4, Google Gemma 3, DeepSeek V3-0324, and Grok 4.1 Fast, respectively.}
\label{fig:non-popular-cultural-subdivision}
\end{figure}

The popular and non-popular title comparisons in Fig.~\ref{fig:popular-cultural-subdivision} and Fig.~\ref{fig:non-popular-cultural-subdivision} show that popular titles generally achieve higher accuracy than non-popular titles. This trend is visible across most languages and categories, suggesting that title popularity affects answer correctness. However, the figures also show some model-level variations. For example, in some non-popular settings, the gap between contexted answering and best RAG becomes larger, indicating that retrieval support is more useful when the baseline performance is weaker.
For RAG-based answering, the overall model ranking remains similar to the preliminary answer-generation results. Meta Llama 4 Maverick 17B remains the strongest model in most settings, while the weaker model varies across languages and categories. The comparison between LangGraph and LlamaIndex shows only marginal differences, with LlamaIndex performing slightly better in some cases. Therefore, the main performance differences are more strongly associated with language, model choice, cultural subdivision, and title popularity than with the specific RAG framework.

\section{Discussion}
\begin{figure}[H]
\centering
\includegraphics[width=\columnwidth]{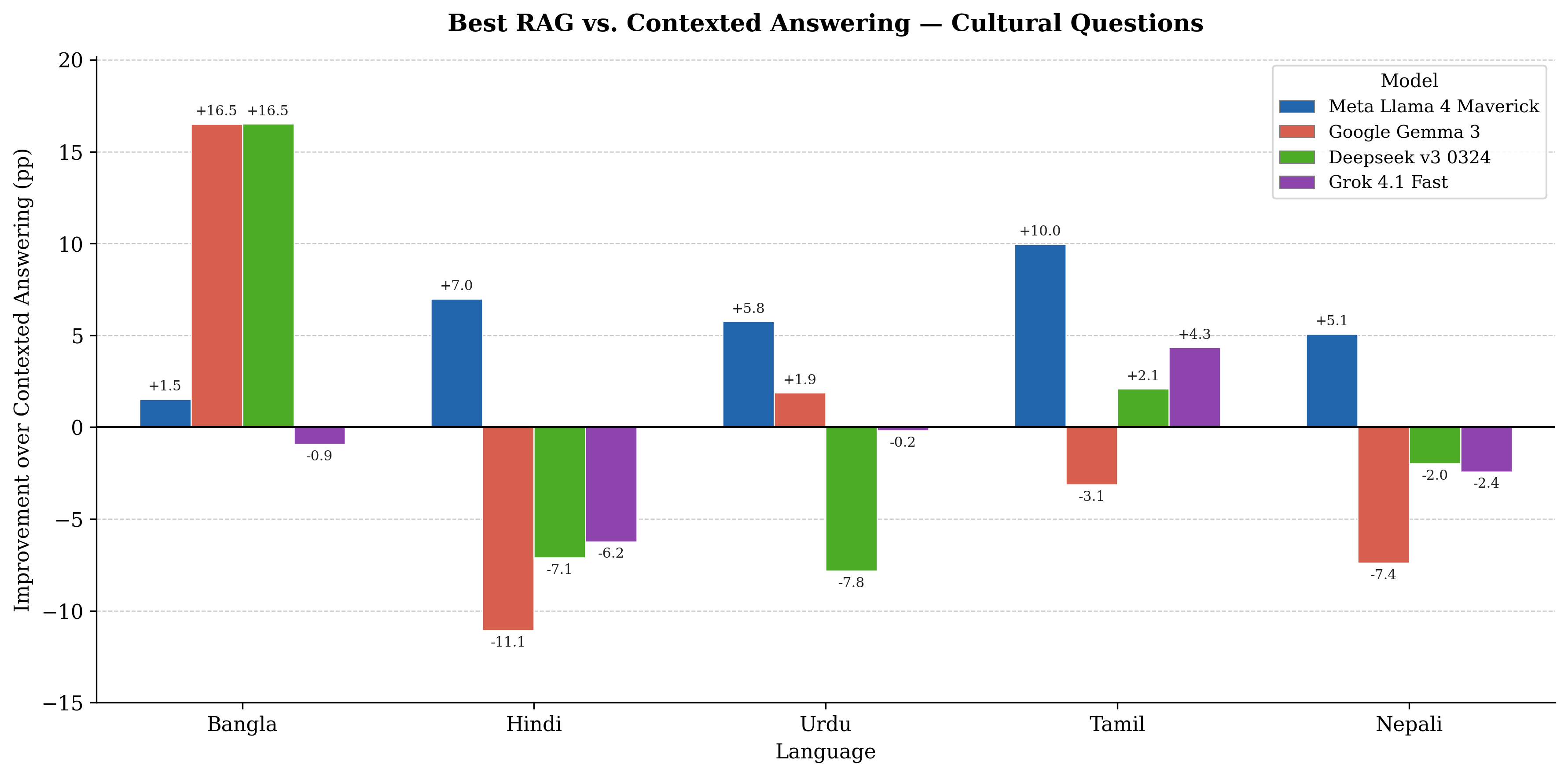}
\caption{Percentage-point improvement of best RAG performance over contexted answering for culturally grounded questions across five languages and four models.}
\label{fig:cultural-rag-improvement}
\end{figure}

The first research question examines whether RAG affects culturally grounded and non-cultural questions differently in the multilingual setup. As shown in Figure~\ref{fig:cultural-rag-improvement}, the effect of RAG on culturally grounded questions varies across languages and models. Bangla and Tamil show the clearest RAG advantage, with most models improving over contexted answering. In contrast, Hindi and Nepali are more context-favored, while Urdu shows a mixed pattern. This indicates that culturally grounded questions do not automatically benefit from retrieval. Instead, the improvement depends on the language, model, and relevance of the retrieved evidence.

\begin{figure}[H]
\centering
\includegraphics[width=\columnwidth]{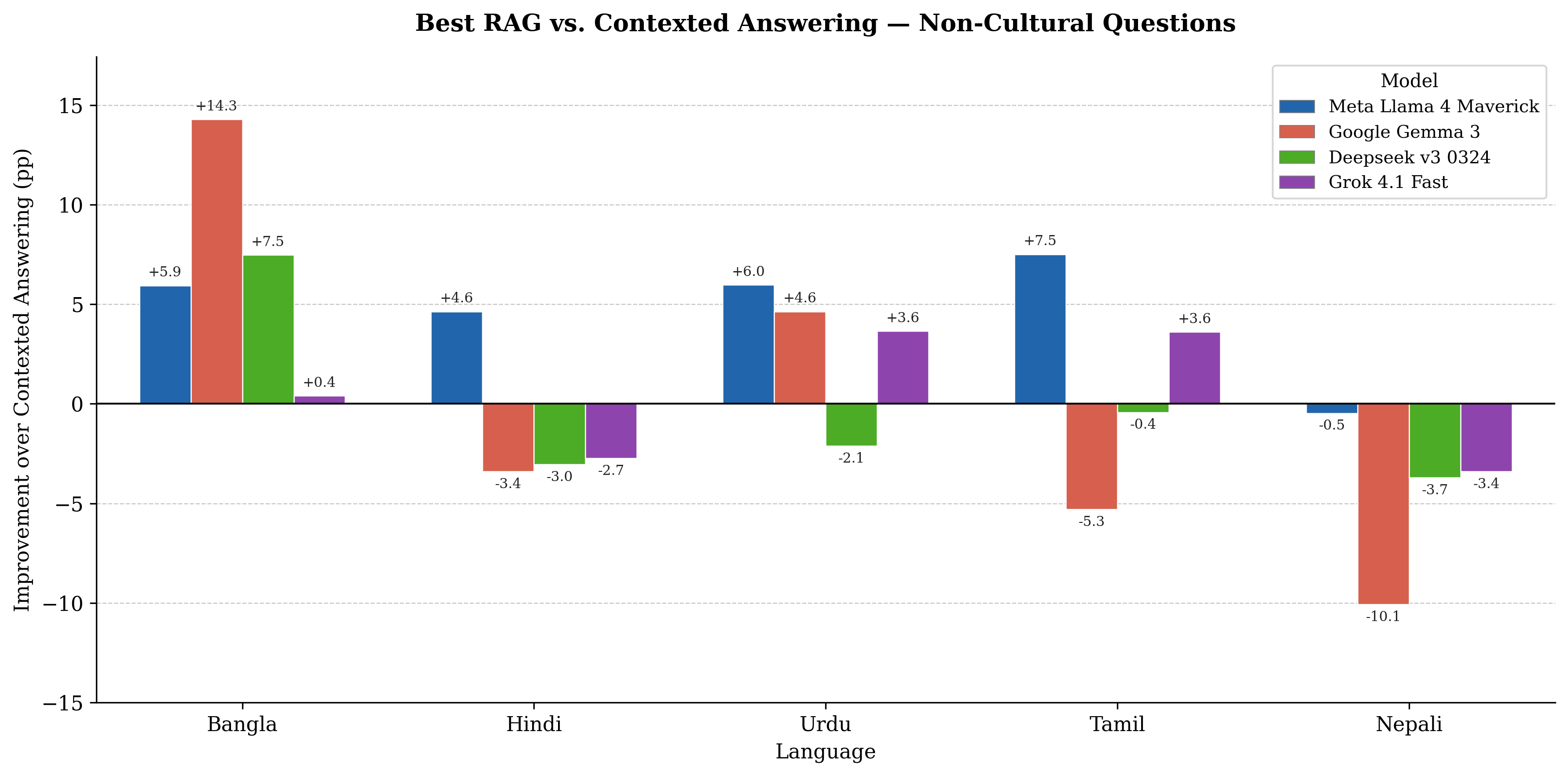}
\caption{Percentage-point improvement of best RAG performance over contexted answering for non-cultural questions across five languages and four models.}
\label{fig:non-cultural-rag-improvement}
\end{figure}

Figure~\ref{fig:non-cultural-rag-improvement} shows that the non-cultural category follows a related but not identical pattern. Bangla shows the most consistent RAG improvement, while Urdu performs better with RAG for most models. However, Hindi and Nepali remain largely context-favored, and Tamil becomes less consistent compared to its cultural-question performance. Therefore, the findings for the first research question suggest that RAG is not simply category-sensitive. Its effect is both language-sensitive and model-sensitive, with stronger cultural-question benefits mainly observed in Bangla and Tamil.

\begin{figure}[H]
\centering
\includegraphics[width=\columnwidth]{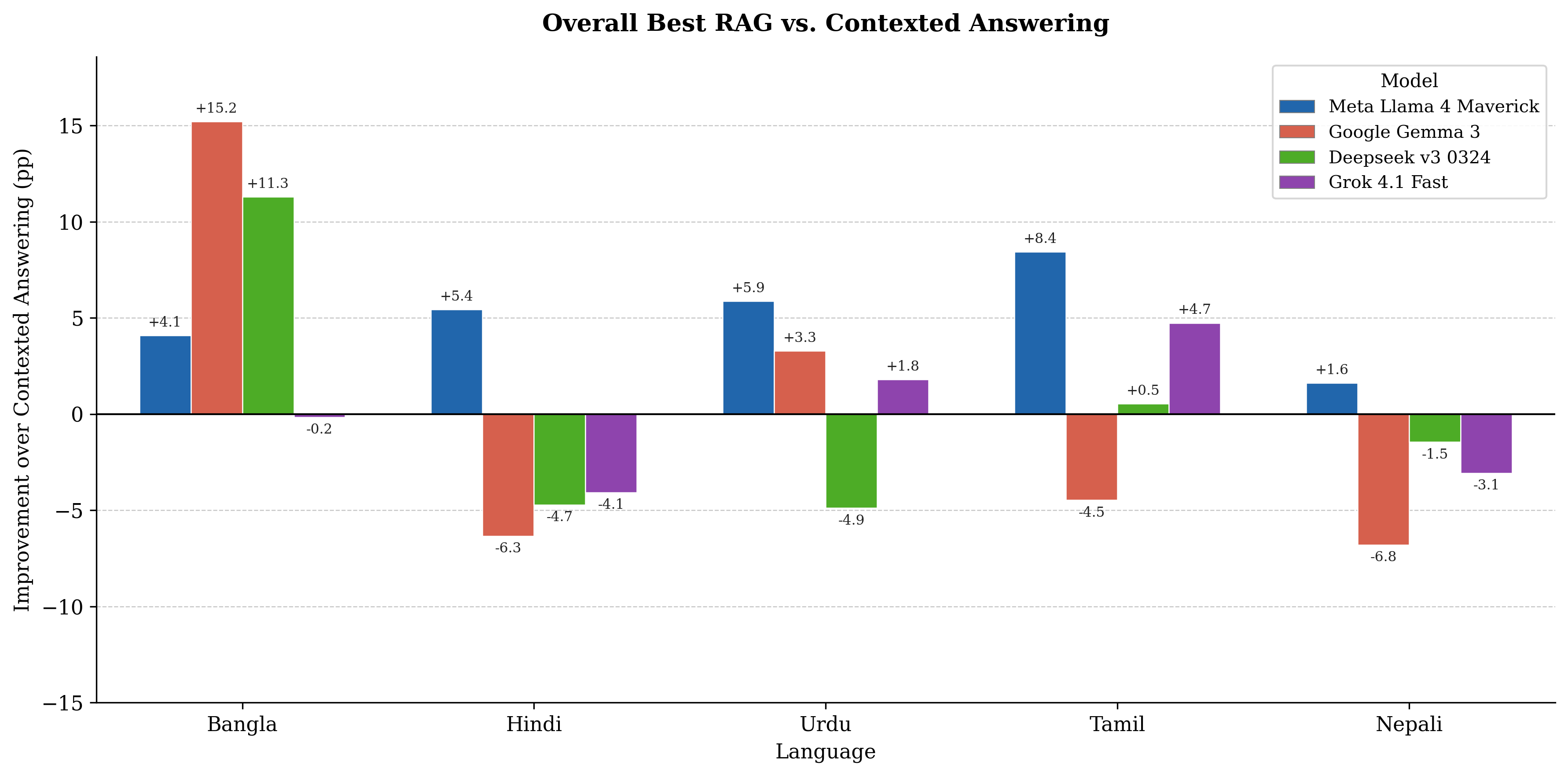}
\caption{Overall accuracy comparison between contexted answering and best RAG performance across five languages.}
\label{fig:overall-rag-improvement}
\end{figure}

The second research question evaluates whether RAG achieves higher overall accuracy than contexted answering. Figure~\ref{fig:overall-rag-improvement} shows that RAG produces higher average accuracy than contexted answering in Bangla, Urdu, and Tamil. However, contexted answering remains stronger in Hindi and Nepali. This means that RAG cannot be treated as an automatically superior method for multilingual QA. Retrieval improves performance when the retrieved passages are useful and well aligned with the question, but it may underperform when the retrieved context is weaker than the directly supplied context.
 
\begin{figure}[H]
\centering
\includegraphics[width=\columnwidth]{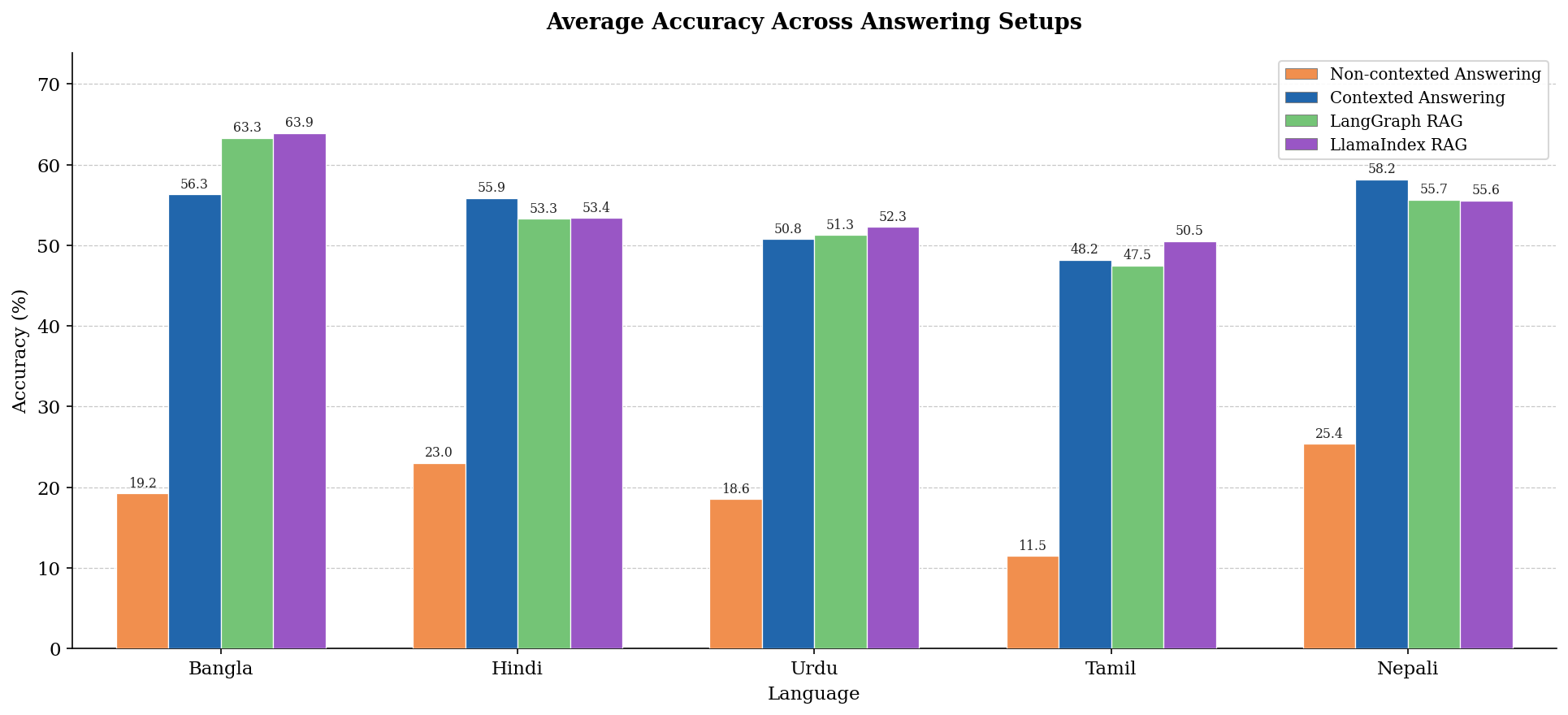}
\caption{Average accuracy across five languages for non-contexted answering, contexted answering, LangGraph RAG, and LlamaIndex RAG.}
\label{fig:avg_accuracy_across_answering}
\end{figure}

The third research question compares the four answering settings: non-contexted, contexted, LangGraph RAG, and LlamaIndex RAG. As shown in Figure~\ref{fig:avg_accuracy_across_answering}, non-contexted answering produces the lowest average accuracy across all five languages. This shows that the models’ internal parametric knowledge alone is not sufficient for reliable answer generation in this South Asian multilingual QA task. Both directly supplied context and retrieval-supported context improve performance substantially. Although prior work has shown that long or excessive context can reduce LLM performance, our results show that, in this setup, the absence of external information is the larger limitation.

\begin{figure}[H]
\centering
\includegraphics[width=\columnwidth]{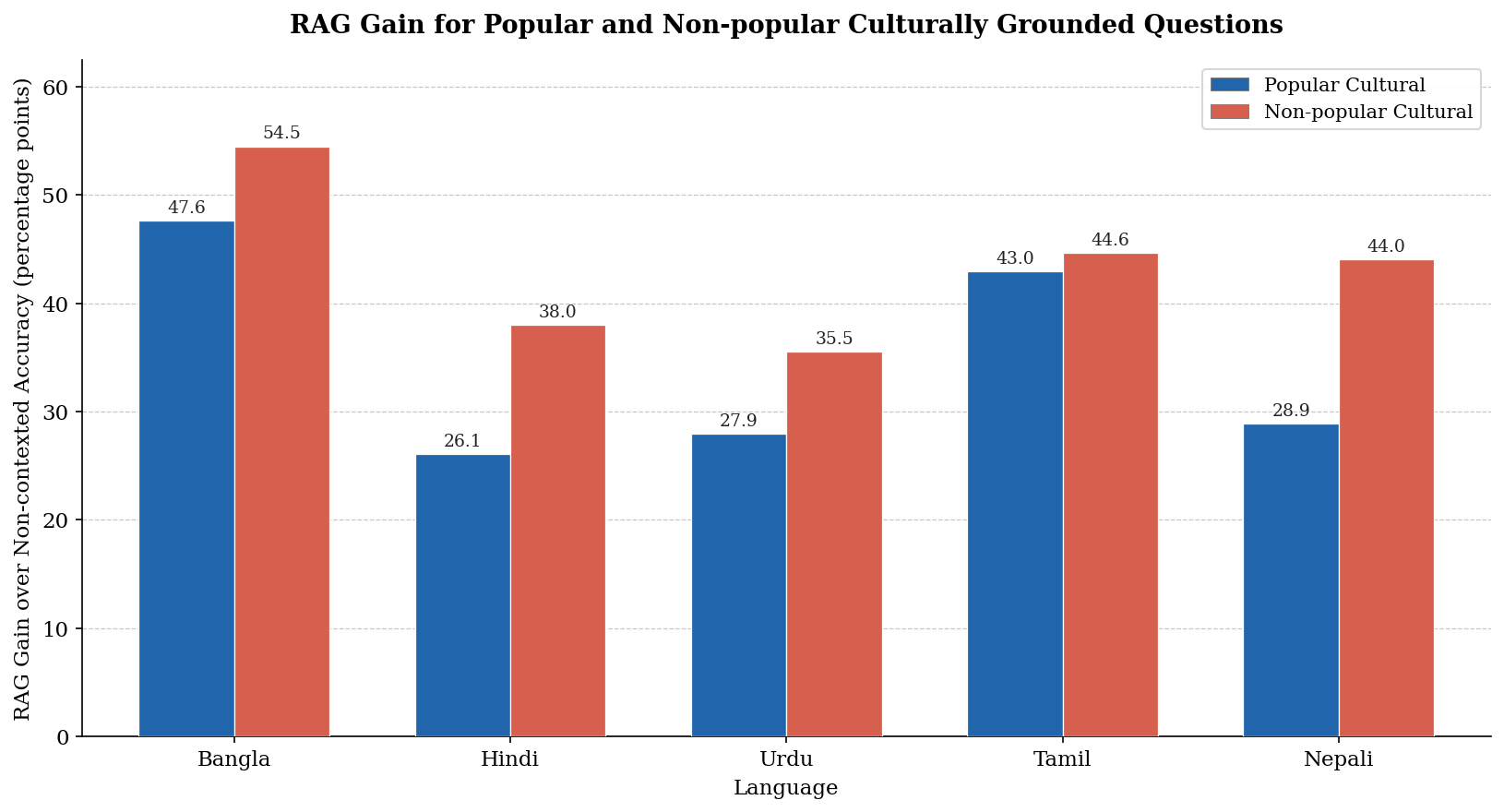}
\caption{RAG gain over non-contexted answering for popular and non-popular culturally grounded questions across five languages.}
\label{fig:RAG_Gain_for_pop_and_non-pop}
\end{figure}

The fourth research question examines whether RAG benefits popular or non-popular culturally grounded questions more. Figure~\ref{fig:RAG_Gain_for_pop_and_non-pop} shows that non-popular culturally grounded questions receive higher RAG gain than popular culturally grounded questions across the evaluated languages. This does not mean that non-popular questions always achieve the highest final accuracy. Rather, it shows that retrieval provides greater relative support when the model starts from a weaker non-contexted baseline. Popular cultural questions appear to be better represented in the models’ internal knowledge, so the additional improvement from retrieval is smaller.
Overall, the findings suggest that RAG should be treated as a context-dependent enhancement rather than a universally superior answering strategy. It reduces knowledge gaps, especially for weaker baseline settings and non-popular cultural questions, but it does not consistently outperform direct contexted answering across all languages. The effectiveness of RAG in multilingual cultural QA depends on retrieval quality, language-specific data alignment, and the ability of each model to use retrieved evidence correctly.

\section{Conclusion}
This study found that LLMs perform better in multilingual cultural knowledge evaluation when supported by retrieved contextual information than being asked to answer from its parametric knowledge base. Earlier experiments have shown that modern LLMs often struggle with culturally nuanced questions when no prior contextual knowledge is provided, especially for less popular or regionally specific languages. LLM-assisted classification helped organize the titles into structured categories, while the later RAG experiments showed that retrieved context could improve answer generation across the multilingual dataset. Overall, the study suggests that multilingual cultural evaluation benefits from combining structured semantic labeling with retrieval-supported generation, especially for short, context-dependent titles and culturally specific questions.

\section*{Acknowledgment}

We express our gratitude and gratefulness to our supervisor, Dr. Farig Yousuf Sadeque for his guidance and support throughout our thesis. We are also grateful to our families and friends who have provided constant encouragement and support.

\end{document}